\documentclass[11pt,a4paper]{article}
\usepackage[hyperref]{acl2021}
\usepackage{times}
\usepackage{booktabs}
\usepackage[ruled,longend]{algorithm2e}
\usepackage{amsmath, amssymb}
\usepackage{mathtools}
\usepackage{latexsym}
\usepackage{todonotes}
\usepackage{hyperref}

\usepackage{graphicx}
\usepackage{subfig}

\DeclareMathOperator{\E}{\mathbb{E}}

\usepackage{enumitem}

\usepackage{microtype}

\usepackage[section]{placeins}

\aclfinalcopy 
\def\aclpaperid{2252} 

\title{Do Multilingual Neural Machine Translation Models\\Contain Language Pair Specific Attention Heads?}

\author{\textbf{Zae Myung Kim\textsuperscript{1,2}, Laurent Besacier\textsuperscript{1}, Vassilina Nikoulina\textsuperscript{1}, Didier Schwab\textsuperscript{2}} \\
  \textsuperscript{1}NAVER LABS Europe\\
  \textsuperscript{2}Univ. Grenoble Alpes, CNRS, LIG\\
  \texttt{\{zae-myung.kim,laurent.besacier,vassilina.nikoulina\}@naverlabs.com}\\
  \texttt{didier.schwab@univ-grenoble-alpes.fr}
}

\date{}

\begin{document}
\setlength{\abovedisplayskip}{0pt}
\setlength{\belowdisplayskip}{0pt}
\setlength{\abovedisplayshortskip}{0pt}
\setlength{\belowdisplayshortskip}{0pt}

\maketitle
\begin{abstract}
Recent studies on the analysis of the multilingual representations focus on identifying whether there is an emergence of language-independent representations, or whether a multilingual model partitions its weights among different languages.
While most of such work has been conducted in a ``black-box'' manner, this paper aims to analyze individual components of a multilingual neural translation (NMT) model.
In particular, we look at the encoder self-attention and encoder-decoder attention heads (in a many-to-one NMT model) that are more specific to the translation of a certain language pair than others by (1) employing metrics that quantify some aspects of the attention weights such as ``variance'' or ``confidence'', and (2) systematically ranking the importance of attention heads with respect to translation quality.
Experimental results show that surprisingly, the set of most important attention heads are very similar across the language pairs and that it is possible to remove nearly one-third of the less important heads without hurting the translation quality greatly.
\end{abstract}

\section{Introduction}

Recent work on analyzing the internals of Transformer-based models \cite{vaswani-etal-1017-attention} sheds some light on how different components within the models affect the final performance \cite{bogoychev2020parameters,behnke-heafield-2020-losing}, and are closely related to playing linguistically interpretable roles \cite{voita-etal-2019-analyzing,jo-myaeng-2020-roles}.
Moreover, studies on the analysis of multilingual representations \cite{conneau-etal-2020-emerging, dufter-schutze-2020-identifying, wang-etal-2020-extending} focus on identifying whether there is an emergence of language-independent representations in multilingual models, or whether multilingual models partition their weights among different languages.

In this paper, we investigate if similar analysis can be made for pretrained multilingual neural machine translation (NMT) models regarding language pair specificity.
More precisely, we analyze multi-head attention in a many-to-one (Transformer-based) NMT model and try to find, through an extensive ablation method on selection of the attention heads, whether some heads are more specific to the translation of a certain language pair than others.

Our contributions are the following: (1) we examine the effectiveness of different attention-based metrics on pruning encoder self-attention and cross attention heads; (2) we find that while it is possible to discover rare heads that are specific to a language pair by using a proposed head selection method, most important heads are language-independent; (3) we also show that around 30\% of heads can be removed with very little loss of performance.

\section{Related Work}

Recent studies analyzed the roles of attention heads in the Transformer models either in language modeling (LM) \cite{Michel-etal-2019-sixteen, clark-etal-2019-bert, jo-myaeng-2020-roles} or NMT \cite{voita-etal-2019-analyzing, behnke-heafield-2020-losing, Michel-etal-2019-sixteen}.
It has been shown that a set of attention heads might be redundant at inference and can be pruned with almost no loss in performance.
In addition, some studies \cite{voita-etal-2019-analyzing, clark-etal-2019-bert} suggested a linguistic interpretation of self-attention heads.
However, most of these analyses were carried out for a single language (in case of LM) or a single language pair (in case of NMT).

In the meantime, efficiency in the cross-lingual transfer of recently released pretrained multilingual language models \cite{devlin-etal-2019-bert, conneau-etal-2020-unsupervised} has boosted an active line of research trying to analyze their representations to understand what favors the emergence of an \textit{interlingua}.
For instance, \citet{pires-etal-2019-multilingual, dufter-schutze-2020-identifying, K2020Cross-Lingual} tried to decouple the effect of shared ``anchors''\footnote{either shared vocabulary or shared special tokens such as \texttt{\big \langle SEP\big \rangle}, \texttt{\big \langle EOS\big \rangle}, etc.} from the rest of the model.
Very recently, \citet{muller2021align} performed a more fine-grained analysis, examining representations at each layer of the model.

Despite the success of massively multilingual NMT models \cite{johnson-etal-2017-googles, bapna-firat-2019-non, aharoni-etal-2019-massively, zhang-etal-2020-improving}, less effort has been made in analyzing multilingual NMT representations.
\citet{kudugunta-etal-2019-investigating} clustered the representations of different languages learned by multilingual NMT models showing that common representations emerge in the encoder.
\citet{marevcek2020multilingual} found that while RNN models (Attention Bridge architecture) \cite{cifka-bojar-2018-bleu, lu-etal-2018-neural}  learn to capture certain linguistic properties with an increasing number of target languages, Transformer models are largely unaffected.
Recent work of \citet{zhang2021share} introduced a conditional routing layer in a form of gate selection between language-specific and language-independent projection, providing some insights on which components allow for the emergence of \textit{interlingua}.

Our  work builds on the findings from the attention heads analyses \cite{voita-etal-2019-analyzing, Michel-etal-2019-sixteen} but attempts to extend them to multilingual NMT, investigating whether it is possible to discover attention heads that are language pair specific.
Also, we experimented with a set of attention-based metrics and analyzed how effective they are in pruning under different language pairs and types of attention.

\section{Methodology}

As our goal was to identify ``important''  attention heads for different language pairs, we first needed to define a metric or a procedure that can capture the notion of ``importance'' of an attention head, and selected heads based on this importance.

In Section~\ref{metrics}, we present a set of metrics that quantify certain aspects of attention weights, which to some extent, can be considered as the importance.
Section~\ref{sbs} illustrates a more direct approach where the importance of a head is defined as the extent of decrease in BLEU scores \cite{papineni-etal-2002-bleu} resulted in pruning the head.

\subsection{Metrics Based on Attention Weights}\label{metrics}
We experimented with three types of metrics that are defined for each attention head, $\mathrm{\tt head}_{l\in L,h\in H}$, where $l$ and $h$ are the indices of layer and multi-head, respectively.
In what follows we define how the metrics were computed for one sentence.
Each metric was computed and averaged over a set of development sentences, then normalized to zero mean and unit standard deviation for ease of comparison.
We note that $|I|$ and $|J|$ were the number of source tokens and/or target tokens, depending on whether we looked at the self-attention of encoder or the encoder-decoder cross attentions.

\paragraph{Confidence} \citet{voita-etal-2019-analyzing} defined the notion of confidence of a head to be the mean of its maximum attention weights, and showed that only a small set of heads are confident and responsible for most of the model's performance.

\begin{align*}
\mathrm{conf}(\mathrm{head}) \coloneqq \frac{1}{|I|} \sum_{i \in I} \max_{j \in J} \alpha_{i,j}
\end{align*}

\paragraph{Variance}
Inspired by \citet{vig-belinkov-2019-analyzing}, we computed the expected position of attention for token $i$ as $\mu_i \coloneqq \E[j|i] = \sum_{j \in J_{}} j \cdot \alpha_{i,j}$, and measured how much each individual position was away from it:\footnote{As we wanted the important heads to have lower variance, we multiplied the score with $-1$.}

\begin{align*}
\mathrm{var}(\mathrm{head}) \coloneqq - \sum_{i \in I} \sum_{j \in J} \alpha_{i,j} \left(\mu_i - j\right)^{2}
\end{align*}

\paragraph{Coverage} \citet{tu-etal-2016-modeling} defined the notion of coverage for encoder-decoder attentions which computes the amount of attention a source token has received.
We extended the idea to the self-attentions in encoder as well.

\begin{align*}
\mathrm{cov}(\mathrm{head}) \coloneqq \sum_{j \in J} \left(\sum_{i \in I} \alpha_{i,j}\right)^{2}
\end{align*}

More details on the metrics are provided in Appendix \ref{Appendix:metrics}.

\subsection{Sequential Backward Selection of Heads}\label{sbs}

\begin{algorithm}
    $\mathrm{selections} \gets \emptyset$\;
    \While{$|\mathrm{selections}| < |L| \cdot |H|$}{
        $\mathrm{bleuMin} \gets \infty$\;
        $\mathrm{headMin} \gets \emptyset$\;
        \For{$\forall \ \mathrm{head}_{l\in L,h\in H} \not\in \mathrm{selections}$}{
            $\mathrm{masks} \gets \mathrm{selections} \cup \mathrm{head}_{l,h}$\;
            $\mathrm{trans} \gets \mathrm{Translate}(\mathrm{masks})$\;
            $\mathrm{bleuDrop} \gets \mathrm{Evaluate}(\mathrm{trans})$\;
            \If{$\mathrm{bleuDrop} < \mathrm{bleuMin}$}{
                $\mathrm{bleuMin} \gets \mathrm{bleuDrop}$\;
                $\mathrm{headMin} \gets \mathrm{head}_{l,h}$\;
            }
        }
        $\mathrm{selections} \gets \mathrm{selections} \cup \mathrm{headMin}$\;
  }
  \KwRet{$\mathrm{selections}$}\;
  \caption{SBS for Head Selection}
  \label{alg:sbs}
\end{algorithm}

Intuitively, a head can be considered as important if its removal results in a drastic decrease in the BLEU scores.
As different combinations of heads can affect the performance differently, we followed the sequential backward selection (SBS) algorithm \cite{Aha1996}, which is a top-down algorithm starting from a feature set of all features (in our case, a set of all heads) and sequentially removing the most irrelevant features that maximize the evaluation metric (in our case, the BLEU score).

The pseudo-code for the head selection procedure is illustrated in Algorithm~\ref{alg:sbs}.
The algorithm first selects a head that, when masked, results in the smallest decrease in the BLEU score; and adds it to $\mathrm{selections}$.
For subsequent iterations, it proceeds similarly, but the $\mathrm{masks}$ now include the heads in $\mathrm{selections}$ as well as the candidate head.
The procedure terminates when all heads are selected.
Note that the time complexity of the algorithm is $\mathcal{O}(|L|^{2}|H|^2)$, where $L$ and $H$ denote the set of layers and attention heads, respectively.
It is a computationally intensive procedure as for each iteration, a test set is translated and evaluated.

\section{Experiments and Results}

\subsection{Preliminary Experiment}
We conducted a preliminary experiment using a many-to-one multilingual model trained on a TED talk dataset \cite{qi-etal-2018-pre}, covering top-20 source languages with the most data.
We observed that patterns of attention heads (measured with the ``confidence'' metric) for both encoder self-attention and encoder-decoder attention were very similar among the language pairs.

For the main experiment, we decided to use a larger and stronger multilingual model for the following reasons: (1) the TED dataset is quite small and the model trained on it achieves lower BLEU scores and may not be regularised very well;
(2) the network capacity of the TED model could be too limited for the language-pair-specific patterns to emerge (if any).
According to a study on BERT's multilinguality \cite{dufter-schutze-2020-identifying}, the increased network capacity (i.e., over-parameterization) is shown to lead to more decoupled representations between languages.

As the multilingual model described in Section~\ref{subsec:exp_settings} is trained on much larger datasets, and has a network capacity larger than the initial TED model while covering fewer language pairs, we expect that the language-pair specificity (if any) is more likely to emerge.

\begin{figure*}[h]
  \includegraphics[width=\textwidth]{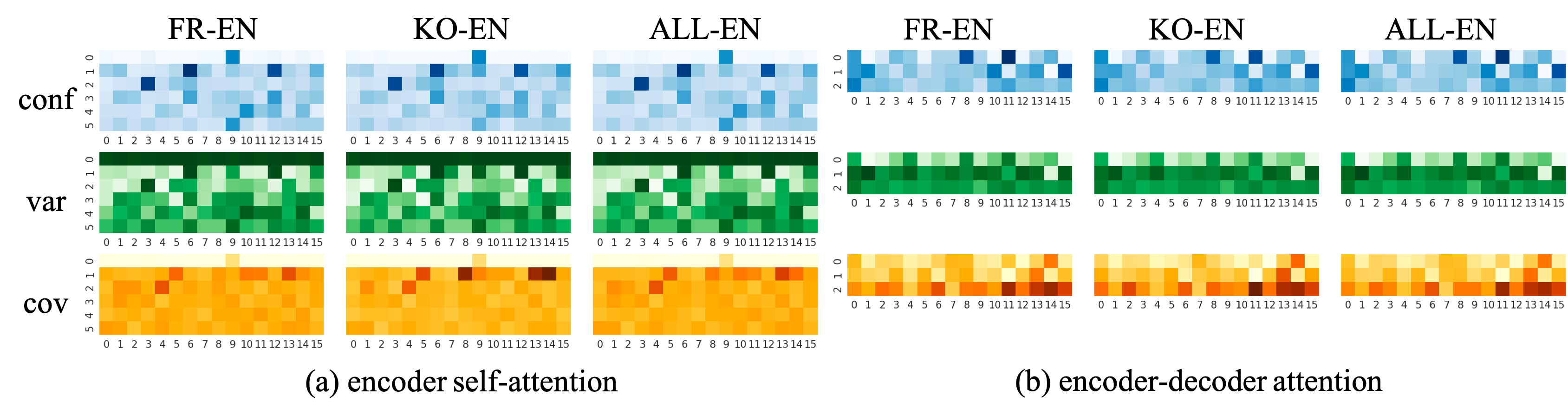}
  \caption{Each heatmap of a language pair shows the corresponding normalized metric scores for every \textbf{(a) encoder self-attention} and \textbf{(b) encoder-decoder attention} head, broken out by layer (vertical axis) and head (horizontal axis). For each metric, the color scales are identical across language pairs.}
  \label{fig:metrics_heatmap}
\end{figure*}

\subsection{Experimental Settings}\label{subsec:exp_settings}
For the sake of reproducibility, all experiments were conducted using a strong publicly available many-to-one multilingual NMT model released by \citet{berard-etal-2020-multilingual}.
The model can translate French, German, Italian, Spanish, and Korean sentences into English.
It is trained with standard open-accessible datasets, including biomedical corpora where available.
The model uses a variant of the Transformer-Big architecture \cite{vaswani-etal-1017-attention} with a shallower decoder: 16 attention heads, 6 encoder layers, and 3 decoder layers.
The model produces  SOTA- or near-SOTA-level results for news, IWSLT, and biomedical translation tasks.

As the model is many-to-one, we could set up a controlled experiment where the BLEU scores were directly compared among the language pairs. 
We employed the development and test sets from the TED talk dataset, and utilized only the multilingual sentence pairs where both source and reference sentences were present for all five language pairs.\footnote{Note that the English reference sentences were the same across the language pairs.}
After the filtering, the development and test sets contained 1771 and 2137 pairs, respectively.

As we were using a many-to-one model, we conducted experiments on both encoder self-attentions and encoder-decoder attentions.
In our experiments, we did not re-train or fine-tune the model when masking each head, making procedure lighter than other approaches involving the re-training.

\subsection{Results}

\paragraph{Heads importance across languages}

Figure \ref{fig:metrics_heatmap} shows the heatmaps for each importance metric (Sect.~\ref{metrics}) for the self-attention and cross-attention heads, respectively.
The heatmaps were computed for each language pair separately (FR-EN, KO-EN, etc.) or jointly for all pairs (ALL-EN).\footnote{We only display FR-EN and KO-EN, reader should refer to  Appendix \ref{Appendix:heatmaps} for all language pairs.}
The main finding is that even if each metric displayed a different heatmap, the important heads were the same for all language pairs according to these metrics.
In other words, the metrics did not highlight the emergence of language pair specific (encoder or cross) attention heads.
Comparing among the metrics, \textit{variance} and \textit{confidence} tended to emphasize the same heads (with the exception of the first self-attention heads of each layer which were systematically rated as important by the \textit{variance} metric).
On the other hand, \textit{coverage} highlighted different heads compared to the other two metrics.

\paragraph{Impact of head selection on NMT performance} In the previous paragraph, we explored several metrics that could help capturing the importance of an attention head.
We now analyze if these metrics could be used to prune heads and the corresponding impact on MT performance.
We also investigate a more direct (but more costly) approach to measure how heads contribute to MT performance.

\begin{figure*}[t]
  \includegraphics[width=\textwidth]{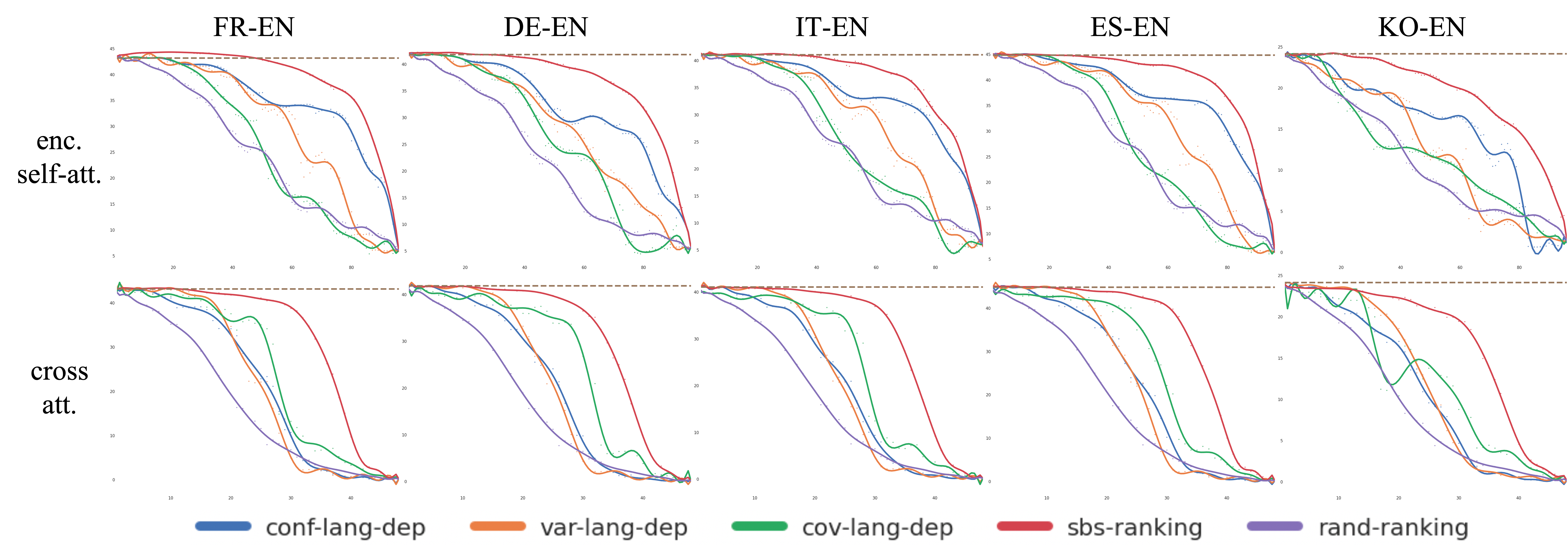}
  \caption{BLEU curves on test set when pruning subsequent \textbf{self-attention} and \textbf{cross attention} heads based on different importance metrics (or SBS) computed from dev set (language pair dependently). To be seen in color.}
  \label{fig:bleu_curve}
\end{figure*}

Figure \ref{fig:bleu_curve} shows the evolution of BLEU curves as more and more heads were pruned.
Head pruning was based on the importance metrics (removing least important heads first according to the metrics presented in Sect. \ref{metrics}) or on the SBS algorithm (Sect. \ref{sbs}).
Head selection was conducted separately for each language pair,\footnote{The language-independent selection of heads led to a very similar plot as Fig. \ref{fig:bleu_curve} and is provided in  Appendix \ref{Appendix:bleu-curves}} and the curves were drawn from fitting polynomial regressions.
First, we observed that, for both encoder self-attentions and cross attentions, it was possible to remove around 30\% of the less important heads without much decrease of BLEU. 
Next, we noted that for cross attention head pruning, \textit{coverage} seemed to be a better alternative than \textit{confidence} and \textit{variance}, while for encoder self-attention pruning \textit{confidence} remained the most efficient.
Intuitively, \textit{coverage} metric is complementary to \textit{confidence} in case of cross attention as it measures whether the whole input has been attended to.
On the other hand, self-attention heads seemed devoted to specific phenomena \cite{voita-etal-2019-analyzing, clark-etal-2019-bert} and there was no need to attend to the whole sentence for this matter.
Finally, we also display the BLEU curves for randomly ranking (\texttt{rand-ranking}) the attention heads, confirming that the metrics proposed can be used as a proxy to measure the importance of heads and prune the least important ones.
However, the exhaustive (but costly) SBS algorithm logically led to the best results.

\begin{figure*}[th!]
  \includegraphics[width=\textwidth]{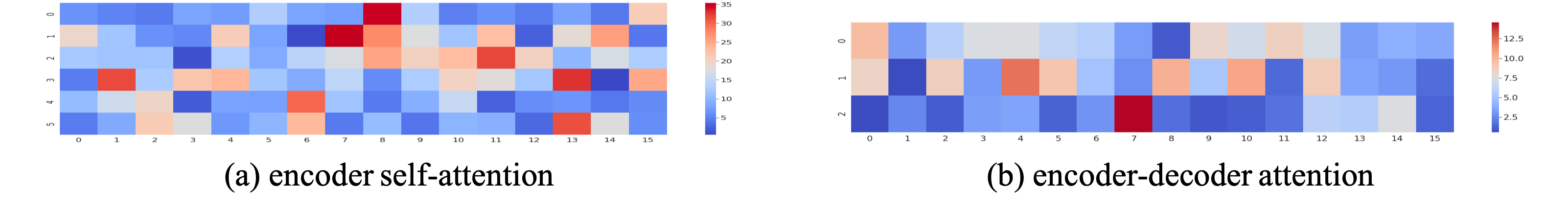}
  \caption{Standard deviation of each \textbf{(a) encoder self-attention} and \textbf{(b) encoder-decoder attention} head ranking with SBS algorithm. SBS rankings range from 1 to 96 for self attention and from 1 to 48 for cross attention and standard deviation is calculated for each head, using these scores, among the five language pairs.}
  \label{fig:sbs_head_std}
\end{figure*}

\paragraph{Is there really no emergence of language-specific heads?} We verified how statistically significant the BLEU differences were between \textit{language-specific} and \textit{language-independent} heads selection processes according to various metrics with Mann–Whitney U tests \cite{mann1947test}.\footnote{We report p-values for Mann–Whitney U tests in the Appendix \ref{Appendix:p-values}.}
We found no significant difference between language-specific and language-independent head rankings, even if some differences emerged for results obtained by SBS ranking.

Finally, we looked at how the individual head rankings were varied according to the SBS algorithm.
Figure~\ref{fig:sbs_head_std} illustrates the standard deviation of each head position among the rankings of the five different language pairs. 
We observed that there were a few heads whose relative importance varied greatly among the language pairs.
For example, the $\mathrm{\tt head}_{2,7}$ of the encoder-decoder attention was ranked as least important for KO-EN but quite important for the other four language pairs.\footnote{Masking this single head alone, resulted in an increase in BLEU for KO-EN by 0.03, while for others, a decrease in BLEU up to 0.5.}
Similarly, $\mathrm{\tt head}_{1,7}$ for encoder self-attention was ranked as not important for ES-EN while very important for KO-EN. 
This analysis showed that, even though the majority of important heads seemed to be language-independent, certain heads may capture different linguistic phenomena.

\section{Conclusion and Future Work}
We investigated if there are attention heads that are language pair specific within a many-to-one multilingual NMT model.
We examined different metrics for heads selection process and found that \textit{confidence} is a good proxy for self-attention heads ``prunability'', and \textit{coverage} is a better indicator for cross attention heads ``prunability''.

We showed that, although it is possible to find the rare heads specific to a language pair via the extensive SBS procedure, the most important heads are language-independent; and it is possible to prune around 30\% of the heads with no retraining and almost no loss in BLEU.\footnote{It is possible that fine-tuning the model after pruning the heads may lead to better BLEU scores, and therefore more aggressive pruning \cite{behnke-heafield-2020-losing,wang-etal-2020-sparsity}.}

As the findings from the SBS procedure indicated that some language pair specific heads do exist, a promising future direction is to perform pruning at different level of granularity \cite{frankle2018the, zhao-etal-2020-masking} (as opposed to single scalar values computed by the metrics) in order to identify which part of the model is more language-specific.
Such analysis could help us to deploy multilingual models with better efficiency / performance trade-offs.

\section*{Acknowledgments}
This work was done as part of the Multidisciplinary Institute in Artificial Intelligence MIAI@Grenoble-Alpes (ANR-19-P3IA-0003).
Authors would like to thank Jaesong Lee and Hyun Chang Cho from NAVER Corp., Kwang Hee Lee from KAIST for insightful comments on the experiments.
We would also like to extend our gratitude to the anonymous reviewers for their valuable feedback.

\bibliographystyle{acl_natbib}
\bibliography{acl2021}

\appendix

\onecolumn

\renewcommand\thefigure{\thesection}
\renewcommand\thetable{\thesection}

\section{Experimental Details}\label{app:exp_details}
All the experiments were conducted using PyTorch \cite{paszke2019pytorch} and Fairseq \cite{ott2019fairseq} toolkit.
The multilingual NMT model used in the experiments can be downloaded online.\footnote{\url{https://github.com/naver/covid19-nmt}}
Please refer to \citet{berard-etal-2020-multilingual} for more details on the model.

When running an experiment for each language pair, a single V100 GPU was used.
We note that computing the SBS rankings for encoder self-attention was the most computationally intensive part, where almost $96^{2}$ translations of the development set were conducted.

When computing the BLEU curves for the \texttt{rand-ranking}, we ran the procedure with a randomly created ranking five times, and averaged the resulting BLEU scores.

\section{Heatmaps of Metric Scores for All Language Pairs}
\label{Appendix:heatmaps}

Figure \ref{fig:heatmaps_for_all_langs} illustrates the normalized metrics scores for every attention head.
We observed that for each metric, the patterns are consistent across all language pairs.

\begin{figure*}[htp]
\subfloat[encoder self-attention]{%
  \includegraphics[width=\textwidth]{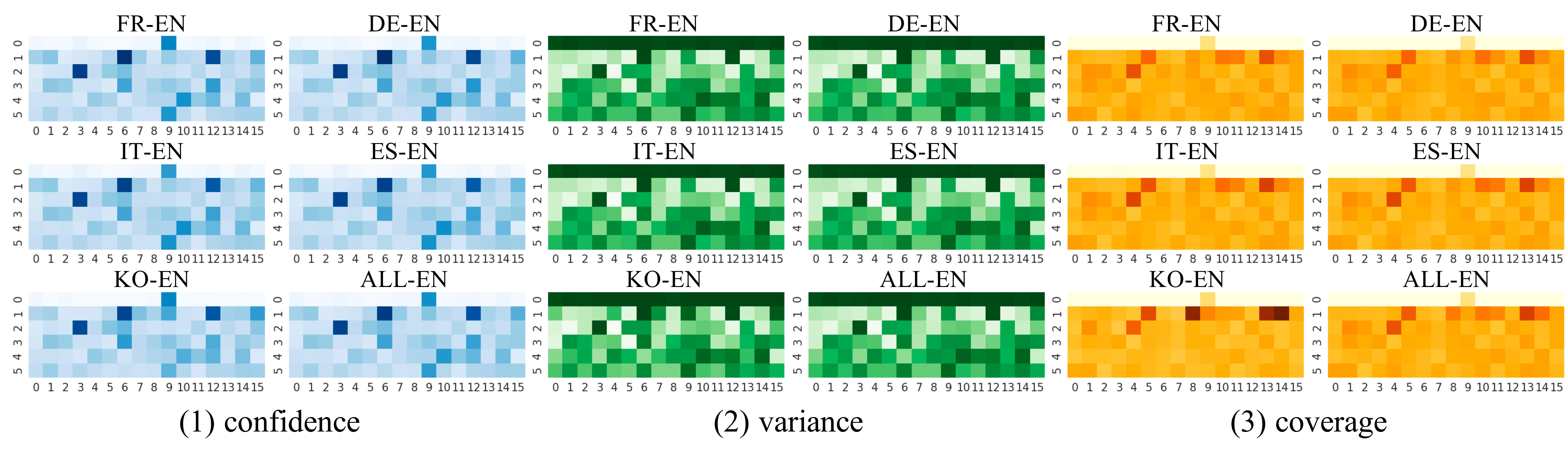}
}\\
\ContinuedFloat
\subfloat[encoder-decoder attention]{%
  \includegraphics[width=\textwidth]{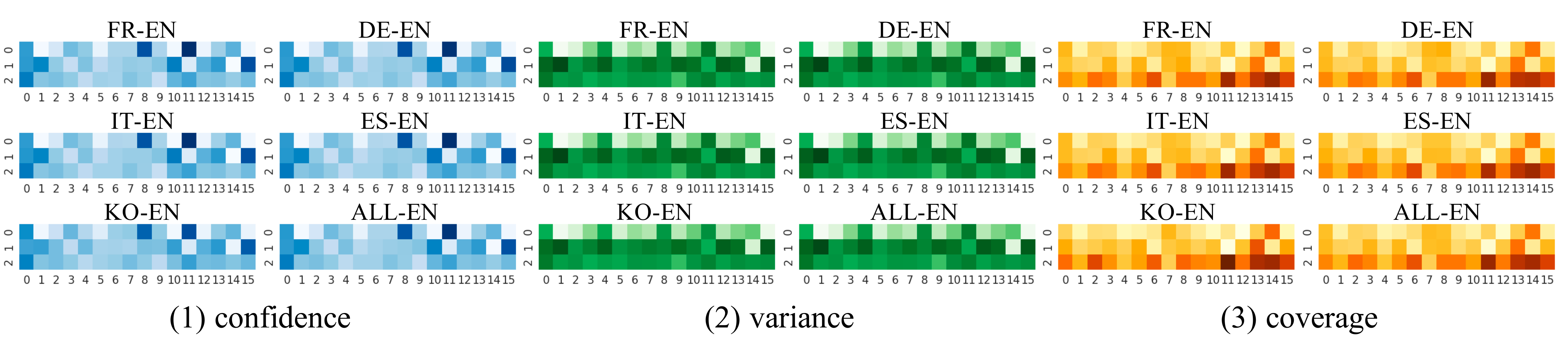}
}
\caption{Each heatmap of a language pair shows the corresponding normalized metric scores for every \textbf{(a) encoder self-attention head} and \textbf{(b) encoder-decoder attention head}, broken out by layer (vertical axis) and head (horizontal axis). For each metric, the color scales are identical across language pairs.}
\label{fig:heatmaps_for_all_langs}
\end{figure*}

\section{Remarks on the Importance Metrics}
\label{Appendix:metrics}

We denote $|L|$ and $|H|$ to be the number of layers and heads, respectively, while $|S|$ and $|T|$ represent the number of source and target tokens.
When calculating each metric, we began with the tensor shape, $(|L|, |H|, |S|, |S|)$ or $(|L|, |H|, |T|, |S|)$, depending on whether we were computing for the encoder self-attention or the cross attention.
After the computation, the shape of the outcome tensor was: $(|L|, |H|)$.

\subsection{Confidence}
We noted that the patterns of the confidence scores for each head tended to vary depending on the length of sentences we used to compute the scores.
This was due to the fact that the metric was calculated by averaging over the maximum attention, which was inversely proportional to the length of sentences.

\subsection{Variance}
The variance metric was defined so that heads with a small variance were considered to be important.
A small variance was achieved when most of the attention weights were focused on one or a few positions.
While this intuition came initially from encoder-decoder attention (interpreting attention as a source-target alignment), it is less clear if it holds for encoder self-attention as well (our results seemed to suggest that it is not the case).

\subsection{Coverage}
While the notion of coverage was initially proposed for encoder-decoder attention, we extended it to the encoder self-attention.
We may consider it as how much a source token has been attended from its neighbouring source tokens.
Similar to the variance metric, for the encoder self-attention, the importance of high coverage is less clear where a head may play a specific role as discussed in \citet{voita-etal-2019-analyzing, clark-etal-2019-bert}.
This probably accounts for the reason that the head pruning of the encoder self-attention was not as effective as that of the cross attention.

\section{P-Values for Mann–Whitney U tests}
\label{Appendix:p-values}

As the BLEU curves obtained from language-specific pruning and language-independent pruning were very similar, we performed a non-parametric statistical test, namely, Mann-Whitney U test, to compare the outcomes.
The test checks whether two samples are likely to derive from the same population (i.e. that the two populations have the same shape).

Table \ref{table:lang_dep_indep_mwu} shows the p-values for the two-sided tests between BLEU curves computed using language-specific and language-independent metrics for encoder self-attention and cross attention.

The high p-values ($> 0.05$) across all language pairs suggest that the differences in the BLEU scores computed from the two scenarios were statistically insignificant.

\begin{table}[!h]

\subfloat[encoder self-attention]{
\centering
\resizebox{0.5\columnwidth}{!}{
\begin{tabular}{lccccc}
\toprule
{} &  FR-EN &  DE-EN &  IT-EN &  ES-EN &  KO-EN \\
\midrule
conf &  0.938 &  0.849 &  0.871 &  0.878 &  0.902 \\
var  &  0.927 &  0.995 &  0.959 &  0.939 &  0.573 \\
cov  &  0.570 &  0.555 &  0.865 &  0.927 &  0.850 \\
sbs  &  0.137 &  0.189 &  0.293 &  0.878 &  0.375 \\
\bottomrule
\end{tabular}}}
\subfloat[encoder-decoder attention]{
\centering
\resizebox{0.5\columnwidth}{!}{
\begin{tabular}{lccccc}
\toprule
{} &  FR-EN &  DE-EN &  IT-EN &  ES-EN &  KO-EN \\
\midrule
conf &  0.918 &  0.988 &  0.965 &  0.968 &  0.881 \\
var  &  0.991 &  0.994 &  0.997 &  0.985 &  0.936 \\
cov  &  0.772 &  0.907 &  0.912 &  0.889 &  0.621 \\
sbs  &  0.901 &  0.631 &  0.936 &  0.918 &  0.404 \\
\bottomrule
\end{tabular}}}
\caption{P-values for Mann–Whitney U tests between BLEU scores computed using language-specific and -independent metrics for \textbf{(a) encoder self-attention} and \textbf{(b) encoder-decoder attention}.}
\label{table:lang_dep_indep_mwu}
\end{table}

\section{BLEU Curves (Language-Independent Head Selection) for All Language Pairs}\label{Appendix:bleu-curves}
\begin{figure*}[htp]
\centering
\subfloat[encoder self-attention]{%
  \includegraphics[clip,width=\textwidth]{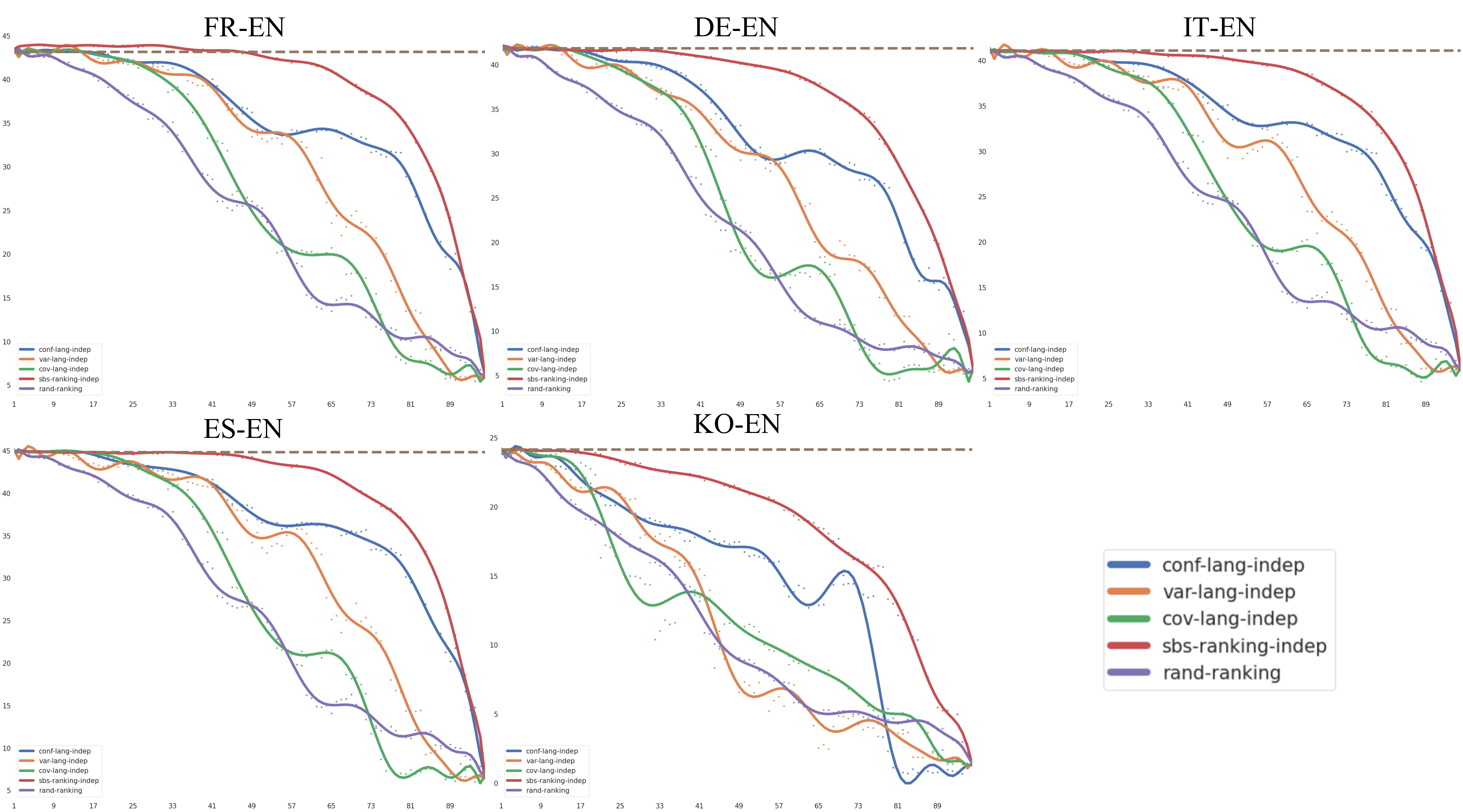}
}\\
\subfloat[encoder-decoder attention]{%
  \includegraphics[clip,width=\textwidth]{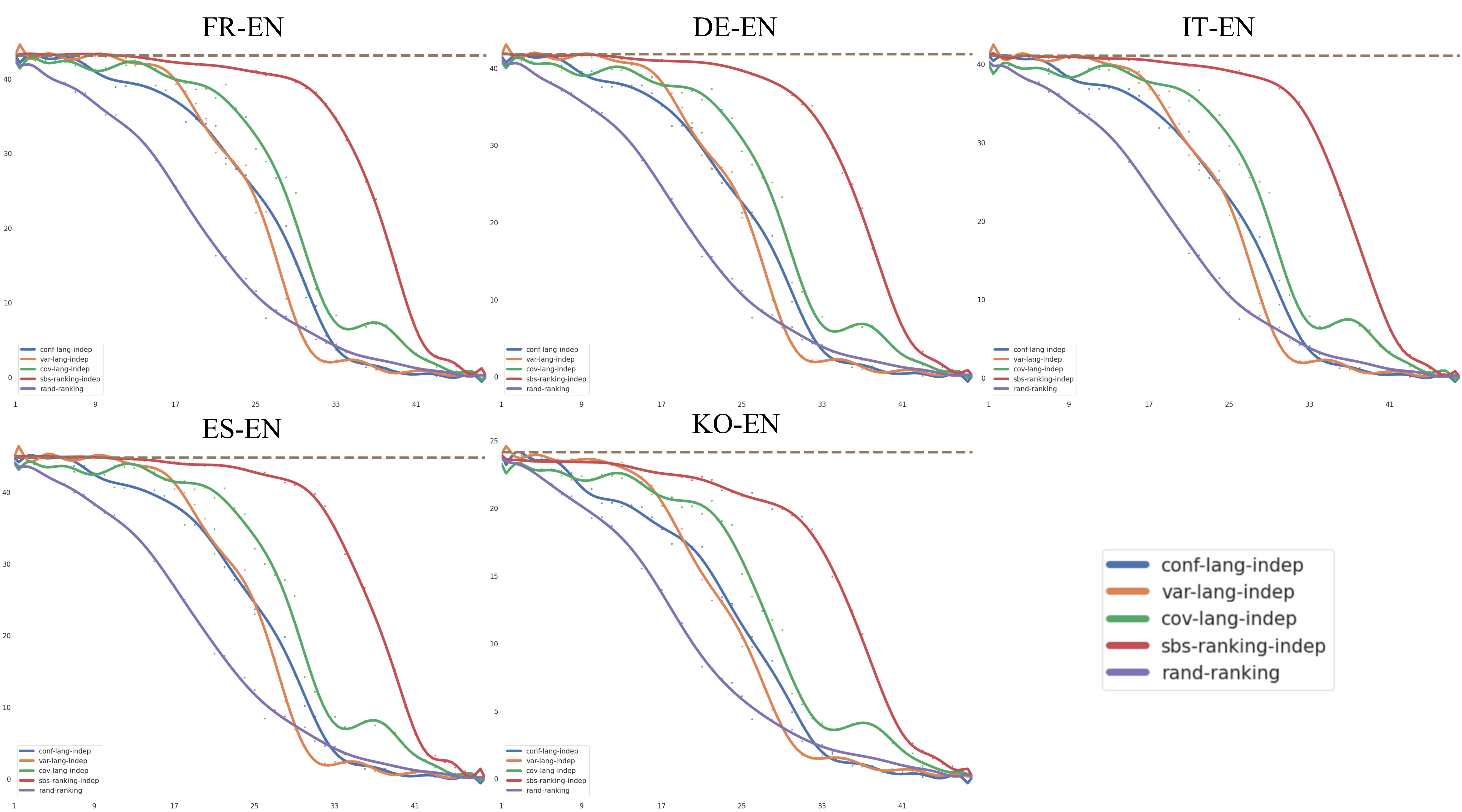}
}
\caption{BLEU curves on test set when pruning subsequent \textbf{(a) encoder self-attention heads} and \textbf{(b) encoder-decoder attention heads} based on different importance metrics (or SBS) computed from the development set (language pair independently).}
\label{fig:bleu_curve_lang_indep}
\end{figure*}

In Figure \ref{fig:bleu_curve_lang_indep}, we present the BLEU curves obtained from pruning the encoder self-attention heads and cross attention heads according to the importance metrics (and SBS) computed over \textit{all language pairs} (i.e. language-independent).
We observed that the curves were very similar to those presented in Figure~\ref{fig:bleu_curve} of the main paper, where the computation was conducted over the \textit{specific language pairs}.

\vspace{-3.5mm}

\end{document}